\documentclass[runningheads]{llncs}
\usepackage[T1]{fontenc}
\usepackage{graphicx}
\usepackage{subcaption}
\usepackage{marvosym}
\usepackage{amsmath}

\usepackage[shortlabels]{enumitem}
\usepackage{tikz}
\usepackage{booktabs}
\usepackage{comment}
\usepackage{amsfonts}
\usepackage{gensymb}
\usepackage{url}
\begin{document}
\title{JAXbounce: A Continuous-Control Pong-Like Benchmark for Active Inference Beyond Low Dimensional Environments}
\titlerunning{JAXbounce}
%
\author{Chaitanya Kaul\inst{1}(\Letter)\orcidID{0000-0003-4893-6222} \and
Markus Klar\inst{1}\orcidID{0000-0003-2445-152X} \and
Sebastian Stein\inst{1}\orcidID{0000-0003-1828-4008} \and
Fraser C. Paterson\inst{1}\orcidID{0009-0003-4392-7092} \and
John H. Williamson\inst{1}\orcidID{0000-0001-8085-7853} \and
Roderick Murray-Smith\inst{1}(\Letter)\orcidID{0000-0003-4228-7962}}

\authorrunning{C. Kaul et al.}
%
\institute{University of Glasgow, Glasgow, Scotland, UK \\
            \email{Chaitanya.Kaul@glasgow.ac.uk}}
%
\maketitle              
\begin{abstract}
We introduce \texttt{JAXbounce}, a continuous control pong-like benchmark designed to evaluate active inference beyond discrete, low dimensional environments. \texttt{JAXbounce} combines continuous states and stochastic dynamics in discrete time with flexible and composable observation models that allow representing the environment as a state vector or a pixel grid, corruptible through the rich family of noise models we provide. It is fully differentiable and parallelisable in JAX. We run over $10,000$ agent simulations on our environment comparing a baseline deterministic solution, to a state-based active inference agent with Unscented Kalman Filter perception and expected free energy planning, and a pixel based agent whose measurement model is a pretrained neural network encoder. All agents are evaluated across three scenarios of increasing difficulty with eight different noise conditions. Our experiments characterise how performance of continuous state agents scales with planning, planning horizons, observation modalities and knowledge of the noise process, exhibiting preference-oriented behaviour even in the presence of increasing uncertainty. We release \texttt{JAXbounce} as a reproducible benchmark for continuous control active inference research. 

\end{abstract}
\section{Introduction}

Active inference combines perception, planning and action into a single unified objective of minimising the expected free energy of candidate sequences of future actions over a finite planning horizon \cite{parr2022active}. The promise of active inference is a single normative objective from which perception, planning, and action follow jointly, and in which exploration is not explicitly added, but is a structural consequence of minimising the expected free energy. An agent's preferences enter the framework as priors over observations rather than an externally specified reward, and information-seeking behaviour arises directly from the epistemic component of the planning objective \cite{friston2016active}. Empirically, these agents are implemented as partially observable Markov Decision Processes (POMDP) acting on low dimensional discrete environments with discrete observation and action spaces \cite{da2020active}\cite{smith2022step}\cite{heins2022pymdp}. Continuous space formulations exist but are run on very low dimensional proof of concept environments \cite{stein2024towards}\cite{waade2021inferring}. With recent interest in continuous control research \cite{klar2025active}\cite{klar2026intermittent}, there is a need for a robust, fully differentiable environment which provides a flexible benchmark to evaluate these agents.

We introduce \texttt{JAXbounce}, a pong-like benchmark environment, to address this gap. Previous research has demonstrated that cultural neurons grown in a dish learn intentional behaviour in the absence of rewards through playing pong \cite{kagan2022vitro}. Learning to play (Atari) pong has also previously been demonstrated by neural networks and deep reinforcement learning \cite{mnih2015human}\cite{schrittwieser2020mastering}. In the context of active inference, playing pong-like games by minimising the expected free energy has been recently studied by simulating \textit{in vitro} cell cultures \cite{friston2025active}\cite{khajehnejad2024biological}\cite{paul2025simulating}. The environments the agents were tested on however were discrete state and primarily low dimensional $5 \times 6$ and $8 \times 4$ grids that do not truly demonstrate their true performance. Existing continuous-control benchmarks \cite{tassa2018deepmind} are well suited for evaluating reinforcement learning algorithms, but do not directly test several properties central to active inference, including uncertainty-aware planning, preference-based behaviour, epistemic exploration, flexible observation models, and scalable belief-state inference. There is currently a research gap in continuous state active inference primarily due to the lack of a unifying benchmark to assess its performance. Towards this, \texttt{JAXbounce} provides continuous states, stochastic dynamics and is fully differentiable and parallelisable in JAX. It contains a rich family of noise models to test continuous state agents while providing the flexibility of tuning the underlying dynamics of the system to evaluate an agent's planning under increasing uncertainty. We evaluate \texttt{JAXbounce} on a series of benchmarks of increasing difficulty. Our contributions are summarised below, 
\begin{itemize}
    \item \textbf{\texttt{JAXbounce.}} We provide a continuous control benchmark for active inference which includes a pong-like environment, as well as a continuous active inference formulation of the game capable of probabilistically inferring an agent's actions from a falling ball's trajectory.
    \item \textbf{Evaluation.} We evaluate our benchmark on various noise conditions, plans, planning horizons and across different scenarios simulating different ball trajectories. We also analyse the ability of the agent to cope with known system dynamics through minimising its epistemic uncertainty of the environment. In total, we evaluate over \textbf{10,000} agents providing a truly robust baseline.
    \item \textbf{Open Source.} We release \texttt{JAXbounce} along with our benchmarks to the community. Code available at https://github.com/drckaul/JAXBounce.
\end{itemize}

\section{JAXbounce}

Most active inference benchmarks have low dimensional state spaces where the difference between raw pixels and observed/belief state is small, leading to little room to study how an inference algorithm degrades as the observation channel deteriorates. We propose an environment implemented in JAX for just-in-time compilation and fast evaluation where an agent must predict a fast-moving target several steps into the future, with the goal being a single brief contact event between two physical objects, and the observation channel can be either a clean low-dimensional state vector or a $R \times R$ grayscale rendering that can be corrupted with realistic image noise. Pong is the simplest arcade game with these properties. We therefore implemented a pong-like game in JAX in a form tailored for variational inference.

\subsection{State and action}

A single \texttt{JAXbounce} trial begins in time and evolves with $\Delta t = 1/60$s time steps. The world state at step $t$ is
\begin{equation}
\mathbf{x}_t = \bigl[\underbrace{\mathbf{x}^{\mathrm{ball}}_t}_{10\text{-D}},\underbrace{\mathbf{x}^{\mathrm{bat}}_t}_{5\text{-D}}\,\bigr]\in\mathbb{R}^{15}.
\end{equation}

\textbf{The ball.} $\mathbf{x}^{\mathrm{ball}}_t$ stores a fall coordinate $y_\mathrm{f}\in[-1,1]$, a lateral coordinate $y_\ell\in[-1,1]$, their velocities $v_\mathrm{f},v_\ell$, the ball radius $r = 0.06$, and geometrical parameters describing the ball. The ball transition dynamics are defined as, $\mathbf{x}^{\mathrm{ball}}_{t+1} = A_{\mathrm{ball}}\,\mathbf{x}^{\mathrm{ball}}_t$, where $A_{\mathrm{ball}}\in\mathbb{R}^{10\times 10}$ advances position by $\Delta t \cdot \mathbf{v}$. 

\textbf{The bat.} $\mathbf{x}^{\mathrm{bat}}_t$ stores the lateral position $y^{\mathrm{bat}}_t$, the bat's velocity, and its fixed geometry (length $L$ and thickness $T = 0.035$). The bat is anchored at the bottom row and only moves laterally. We empirically estimate and fix $L = 0.0625$ for our experiments. The bat dynamics are defined as, $\mathbf{x}^{\mathrm{bat}}_{t+1} = A_{\mathrm{bat}}\,
\mathbf{x}^{\mathrm{bat}}_t + B_{\mathrm{bat}}\,a_t$, with a damping of $0.95$ The control matrix $B_{\mathrm{bat}}$ is chosen so that the action acts as a normalised target velocity, displacing the bat by $\Delta t\,a_t$ within the step.

\textbf{Action.} An agent issues a scalar command $a_t\in[-1,1]$ at each step which is the normalised target velocity of the bat. Within a timestep, the ball moves under its current velocity, and the bat translates by the commanded velocity based on their respective transition functions.

\subsection{A differentiable collision signal}

A trial is a success when the bat and ball collide. We represent a collision between the bat and the ball with a smooth scalar $c_t\in[0,1]$:
\begin{align}
u^{\mathrm{ov}}_t &= r + \tfrac{T}{2} - |\,y_{\mathrm{f},t} - \tilde y^{\mathrm{bat}}|,\\
l^{\mathrm{ov}}_t &= r + \tfrac{L}{2} - |\,y_{\ell,t} - y^{\mathrm{bat}}_t|,\\
c_t &= \sigma(k\,u^{\mathrm{ov}}_t)\sigma(k\,l^{\mathrm{ov}}_t),\qquad k=8.
\label{eq:c}
\end{align}
Here, $T$ and $L$ are the bat thickness and its length respectively, and $r$ is the ball radius. $\tilde y^{\mathrm{bat}}$ is the fixed fall coordinate of the bat. The product is high only when both axes overlap (vertical denoted by $u^{\mathrm{ov}}_t$ and lateral denoted by $l^{\mathrm{ov}}_t$) and varies smoothly. $k=8$ was tuned so that $c_t$ peaks near $0.5$ at perfect contact for the bat length and falls below $0.1$ at distances greater than the radius of the ball.

\subsection{Observation channels}
\label{sec:obs}

An agent does not see the world state $\mathbf{x}_t$ directly, but through one of two observation functions. The state-based agents observe a 4-D vector
\begin{equation}
\mathbf{o}^{\mathrm{s}}_t = g_{\mathrm{s}}(\mathbf{x}_t) = [y_{\mathrm{f},t}, y_{\ell,t}, y^{\mathrm{bat}}_t, c_t]\in\mathbb{R}^4,
\label{eq:o-deterministic}
\end{equation}
which are the ball's position, the bat's position, and the differentiable collision signal $c_t$ (equation \eqref{eq:c}). Agents with pixel-based inputs observe a grayscale image $\mathbf{o}^{\mathrm{p}}_t = g_{\mathrm{p}}(\mathbf{x}_t)\in[0,1]^{R\times R}$ with the ball rendered as a blob and the bat as a rectangle. We experiment with different values of $R$ to evaluate the effect of different grid sizes on active inference agents.

\subsection{Scenarios and catch criterion}
\label{sec:senarios}

Three different scenarios are designed in increasing order of difficulty. The ball always falls from the top of the grid $y_{\mathrm{f},0}=-0.9$ at the start of each trial with a different location across the x-axis sampled from a uniform distribution $y_{\ell,0}\sim\mathcal{U}[-0.7,0.7]$. In scenario 1, \textbf{S1} (easy), the ball falls vertically with a constant speed. In scenario 2, \textbf{S2} (medium), The ball falls vertically with an unknown speed in each trial and the agents must infer the ball's velocity. In scenario 3, \textbf{S3} (hard), the ball falls at a lateral angle of $\pm 35\degree$ with an unknown speed. A trial within each scenario counts as a success if $\max_t c(\mathbf{x}_t)>0.4$ which equates to a physical overlap between the bat and the ball. 

\section{Active Inference for JAXbounce}
\label{sec:aif}

Active inference treats perception and action as inference under a single objective. Perception is the agent's posterior belief over the latent states of the world given a stream of observations. Action is the choice of control that minimises the expected free energy of its predicted future observations. Our resulting active inference framework consists of the following components,

\begin{itemize}
\item \textbf{Generative process.} The real world \texttt{JAXbounce} dynamics and one of the observation functions $g_{\mathrm{s}}$ or $g_{\mathrm{p}}$. The process is the real world in which the agent acts, observed through its Markov blanket. The agent does not have access to $\mathbf{x}_t$ directly, but only to its (noisy) projection $g(\mathbf{x}_t)$.

\item \textbf{Generative model.} The agent's internal model of how observations are generated from latent states. Each agent type of agent has its own specific measurement model $\mathbf{h}$. For our state-based agents, $\mathbf{h}=g_{\mathrm{s}}$ and for our pixel-based agents, $\mathbf{h}=\phi_\theta(g_{\mathrm{p}})$ where a rendered \texttt{JAXbounce} grid is passed through a pretrained neural network to estimate the observed states. $\phi$ is the pretrained neural network encoder with $\theta$ parameters.

\item \textbf{Probabilistic beliefs.} All agents we run carry two Gaussian beliefs at each simulation step. $Q^{\mathrm{s}}_t := \mathcal{N}(\boldsymbol{\mu}^{\mathrm{s}}_t,\boldsymbol{\Sigma}^{\mathrm{s}}_t)$ over the 15-D latent state, and $Q^{\mathrm{k}}_t := \exp(\mathcal{N}(\boldsymbol{\mu}^{\mathrm{k}}_t,\boldsymbol{\Sigma}^{\mathrm{k}}_t))$ over the observations. Each belief is a mean and covariance matrix of a multivariate Gaussian distribution and they are log-Gaussian to keep their scale positive at all times. 

\item \textbf{Belief update.} To update an agent's beliefs after performing an action, we used an Unscented Kalman Filter (UKF) \cite{wan2000unscented}\cite{julier2004unscented}. The UKF's prediction stage propagates the state belief forward through the system dynamics via the unscented transform (UT). In the correction stage we compare predicted belief against an observation using variational inference (VI). Given smooth system dynamics the unscented transform provides a cheap and exact belief estimate while the VI correction helps account for any non-nonlinearities arising from differentiable measurement models.

\item \textbf{Action selection.} Action selection is done by minimising the expected free energy (EFE) of simulated candidate future action sequences of horizon $H$. The EFE decomposes into a pragmatic and an epistemic term. We primarily experiment with the pragmatic term, minimising EFE to satisfy our set preference, and later show experiments where the epistemic term is useful in cases where unknown system dynamics require planning to reduce uncertainty about the latent states.
\end{itemize}

\subsection{Beliefs and belief update}
\label{sec:perception}

\textbf{Beliefs.} We parameterise the state belief at timestep $t$, $Q^{\mathrm{s}}_t$, by a mean $\boldsymbol{\mu}^{\mathrm{s}}_t\in\mathbb{R}^{15}$ and a lower-triangular Cholesky factor $\mathbf{L}_t$ such that $\boldsymbol{\Sigma}^{\mathrm{s}}_t = \mathbf{L}_t\mathbf{L}_t^{\!\top}$. The Cholesky form keeps the covariance matrix positive-definite by construction throughout VI optimisation.

\textbf{Observation noise belief.} The observation noise belief, $Q^{\mathrm{k}}_t$ enters the VI based correction step as how much an agent trusts each value in its 4D observation vector. Our ablations show how mis-calibrating $Q^{\mathrm{k}}$ by having a strong belief that the agent's measurement model provides noise free observations can result in the agent to over trust the noisy observations.

\textbf{UKF Predict step.} Sigma points are placed at $\boldsymbol{\mu}^{\mathrm{s}}_t$ and at $\boldsymbol{\mu}^{\mathrm{s}}_t\pm[\sqrt{(d+\lambda)\boldsymbol{\Sigma}^{\mathrm{s}}_t}]_j$ for $j=1,\dots,d$, with $d=15$, scaling $\alpha=10^{-3}$, $\beta_{\mathrm{ut}}=2$, $\kappa=3\times10^{6}$. Each of the $2d+1=31$ sigma points is propagated through the system dynamics, $f(\cdot,a_t)$, and the predict-prior $\bar Q^{\mathrm{s}}_{t+1} = \mathcal{N}(\bar{\boldsymbol{\mu}}^{\mathrm{s}}_{t+1},\bar{\boldsymbol{\Sigma}}^{\mathrm{s}}_{t+1})$ is recovered as the weighted mean and covariance of the propagated points.

\textbf{UKF Correct step.} The variational posterior is found by minimising the variational free energy
\begin{equation}
\mathcal{F}(Q^{\mathrm{s}}) = D_{\mathrm{KL}}(Q^{\mathrm{s}}\,\Vert\,\bar Q^{\mathrm{s}}_{t+1}) + \mathbb{E}_{Q^{\mathrm{s}}Q^{\mathrm{k}}}[-\log p(\mathbf{o}_{t+1}\mid\mathbf{x},\boldsymbol{\sigma}_o)],
\label{eq:vfe}
\end{equation}
which trades off staying close to the predict-prior (first term) against fitting the new observation under the measurement model i.e. the log-likelihood of the observation under the measurement model (second term).
$\mathcal{F}$ is minimised by gradient descent on $(\boldsymbol{\mu}^{\mathrm{s}},\mathbf{L})$, with the expectation in equation \eqref{eq:vfe} estimated from samples drawn from $Q^{\mathrm{s}}$ and $Q^{\mathrm{k}}$.

A small ridge term, $\epsilon\mathbf{I}$ ($\epsilon=10^{-8}$), is added to the state covariance before every Cholesky decomposition to keep the belief covariance positive-definite.

\subsection{Action selection: expected free energy}
\label{sec:efe}

The agent's preferences over observations are encoded as a prior preference distribution $P^{\mathrm{c}}$. We encode this with a Gaussian preference on the collision channel of the observation,
\begin{equation}
P^{\mathrm{c}}(\mathbf{o}) = \mathcal{N}(o_c,\sigma_C^{2}),
\label{eq:preference}
\end{equation}
where $o_c = o_4$ is the collision observation. The tightness $\sigma_C^{2}$ controls how strongly the agent insists on the collision signal being close to its target value of $1$.

\textbf{Expected free energy.} The agent scores a candidate plan $\tilde{\mathbf{a}}=(a_{t+1},\dots,a_{t+H})$ by the expected free energy of the observations it expects the plan to produce. Without the epistemic term,
\begin{equation}
\mathcal{G}_p(\tilde{\mathbf{a}}) = -\mathbb{E}_{Q^{\mathrm{o}}(\tilde{\mathbf{a}})}[\log P^{\mathrm{c}}(\mathbf{o})],
\label{eq:efe-pragmatic}
\end{equation}
where $Q^{\mathrm{o}}(\tilde{\mathbf{a}})$ is the agent's predictive distribution of observations under plan $\tilde{\mathbf{a}}$.

\textbf{Hallucinated observations.} In the absence of real observations, to evaluate equation \eqref{eq:efe-pragmatic} the agent simulates the observations the plans would produce. For each timestep in the planning horizon, $\tau \in \{t+1, t+2, ... t+H \}$ the agent runs a UKF prediction step under each action $a_\tau$ in the timestep to obtain its predicted state belief $Q^{\mathrm{s}}_\tau(\tilde{a})$. We then sample $N_s$ states from this predicted belief and $N_o$ observations each to estimate their log-likelihood given the preference distribution. The plan with the lowest expected free energy is chosen as the optimal plan at that timestep by the agent. $N$ plans are rolled out for a planning horizon $\tau$ in this step and the executed plan is the argmax of the per plan negative expected free energy,
\begin{equation}
    \mathrm{NEFE}(\tilde{\mathbf{a}}^{(i)}) = - \beta \mathcal{G}_p(\tilde{\mathbf{a}}^{(i)}),
\end{equation}
where $\beta = 5$ is the pragmatic scaling term. The first action of the plan with the lowest NEFE is chosen and executed.

\subsection{Agents}

To evaluate our active inference framework, we propose three agents. The first is a deterministic baseline that executes actions based on predicting a closed-form solution of the ball's landing point without active inference, the second is a state-based active inference agent which selects actions to minimise the expected free energy of its hallucinated future observations against its preference via a state based measurement model, and the third is a pixel-based active inference agent that takes a \texttt{JAXbounce} grid as an input and uses a pretrained encoder as its measurement model to predict its observations. All three agents share the \texttt{JAXbounce} dynamics, the UKF and the criteria that defines a successful catch. They differ in their measurement model $\mathbf{h}$ and in how they select actions $\mathbf{a_t}$ at each timestep.  

\paragraph{\textbf{Deterministic Baseline.}} The deterministic agent perceives the world through its observation states (Section \ref{sec:obs}). At every step, the deterministic baseline uses the UKF's posterior belief of $\boldsymbol{\mu}^{\mathrm{s}}_t$ to obtain a belief about the ball's position $(\hat{y_\mathrm{f}}, \hat{y_\ell})$, its velocity $(\hat{v_\mathrm{f}}, \hat{v_\ell})$, and the position $\hat{y}^{\mathrm{bat}}$, of the bat. The agent then computes the time for the ball to reach the bat, from which it then calculates its landing point as, 
\begin{equation}
t_{\mathrm{g}}=\max\Bigl(0,\frac{\tilde y^{\mathrm{bat}}-\hat y_\mathrm{f}}{\hat v_\mathrm{f}}\Bigr), \quad \hat y_{\mathrm{land}}=\hat y_\ell+\hat v_\ell\,t_{\mathrm{g}}, \quad a_t=\mathrm{clip}\bigl(k_p(\hat y_{\mathrm{land}}-\hat y^{\mathrm{bat}}_t),\pm 1\bigr),
\label{eq:deterministic}
\end{equation}
Where $\tilde{y}^{\mathrm{bat}}$ is the location of the bat on its fixed axis and $k_p = 2$ is the gain that limits the bat action to within the grid. The gain is fixed to a value $>1$ to allow the deterministic solution to reach the ball even for harder scenarios, making it a competitive baseline for the active inference agent.

\paragraph{\textbf{State-Based Active Inference.}} The state-based AIF agent shares its observation model with the deterministic agent. They differ in their action selection methodology. The state-based AIF agent has a preference for observing a collision between bat and ball. It selects actions towards its preference by minimising the expected free energy of rolled out plans across fixed planning horizons with this preference and executes the first action in the selected plan with the least expected free energy. A comparison with the deterministic agent under this setting allows us to directly see the advantage of active inference based planning in our environment as this tests which agent can handle sensor uncertainty better.

\paragraph{\textbf{Pixel-Based Active Inference.}} The pixel-based AIF agent differs from the state-based AIF agent in as its measurement model. Its action selection is similar to the state-based AIF agent. 
The measurement model is an MLP that takes a \texttt{JAXbounce} grid as an input and outputs the 4 observation channels from Section \ref{sec:obs}. It flattens the rendered frame to a $1D$ vector and applies two ReLU activated hidden layers of width $128$ and $64$ before a linear $4$-dim output $\mathbf{o}^{\mathrm{s}}_t$. We pre-train 4 such encoders, one for each image grid size we test in our experiments. The encoder's pretraining is described in appendix \ref{sec:encoder}.

\section{Experiments}

To evaluate the ability of active inference agents to play JAXBounce, we compare the deterministic baseline solution with the state-based AIF agent. We also run experiments with the pixel-based AIF agent for reference. Each agent in each experiment is run for 30 trials and a successful trial is one where the agent can align the bat with the ball to make contact with it. We run all agents across all 3 scenarios (Section \ref{sec:senarios}) and 8 noise conditions. For the deterministic baseline and state-based AIF agent, the noise conditions are shown in Table \ref{tab:noise-state}, which increase the uncertainty about the ball's initial position, its speed which the agent needs to infer, and uncertainty for the measurement model to understand how the agents handle noisy environments. For the pixel-based AIF agent, we apply gaussian noise ($\sigma = 0.05, \, 0.10, \, 0.20$), Hann-window blur (windows $3 \times 3,\, 5 \times 5$) and salt and pepper noise (per-channel rate $0.02 \, \mathrm{and} \, 0.05$) to the pixel grids as the 8 noise conditions. \\
We evaluate two forms of agents for our experiments. In the first case, the agent has an initial belief about the noise conditions equivalent to Noise 1 in table \ref{tab:noise-state}. This simulates a scenario where the agent assumes the states of the environment are relatively noise free, and we observe how its performance degrades with increasing noise conditions. In the second case, for each simulation, the agent knows the noise in the environment, i.e. its initial belief of the noise is equivalent to the noise condition tested from table \ref{tab:noise-state}. This simulates a scenario where the agent knows the environment is noisy and we test its robustness to this noise through our simulations. Our experiments can be summarised as the following: For case 1, \textbf{Experiment 1 (Bat Size):} Bigger bat sizes make observing collision easier. We experiment with 3 different bat sizes, each making the scenarios harder as the agents now have a lesser margin for error. 
\textbf{Experiment 2 (Plans):} To observe the effect of planning horizons, we run all agents with $100, 200, 300 \,\mathrm{and} \, 400$ plans.
\textbf{Experiment 3 (Planning Horizons):} We experiment with planning horizons of $10, 60 \, \mathrm{and} \, 100$ timesteps to see the effects of allowing agents to hallucinate observations over longer planning horizons to minimise its expected free energy.
\textbf{Experiment 4 (Pixel-based AIF grid-size):} As the \texttt{jaxBounce} grid is a hyperparameter that defines the representation learned by the MLP encoder of the pixel-based AIF agent, we experiment with 3 grid sizes, $16 \times 16, 32 \times 32 \, \mathrm{and} \, 64 \times 64$.
For case 2, \textbf{Experiment 5 (Agent knows the noise conditions):} we evaluate the state AIF agent across $100, 200, 300 \,\mathrm{and} \, 400$ randomly sampled plans for horizons of $10, 60 \, \mathrm{and} \, 100$ timesteps.

\section{Results}

\begin{figure}[htbp]
 \centering
 \includegraphics[width=1.0\textwidth]{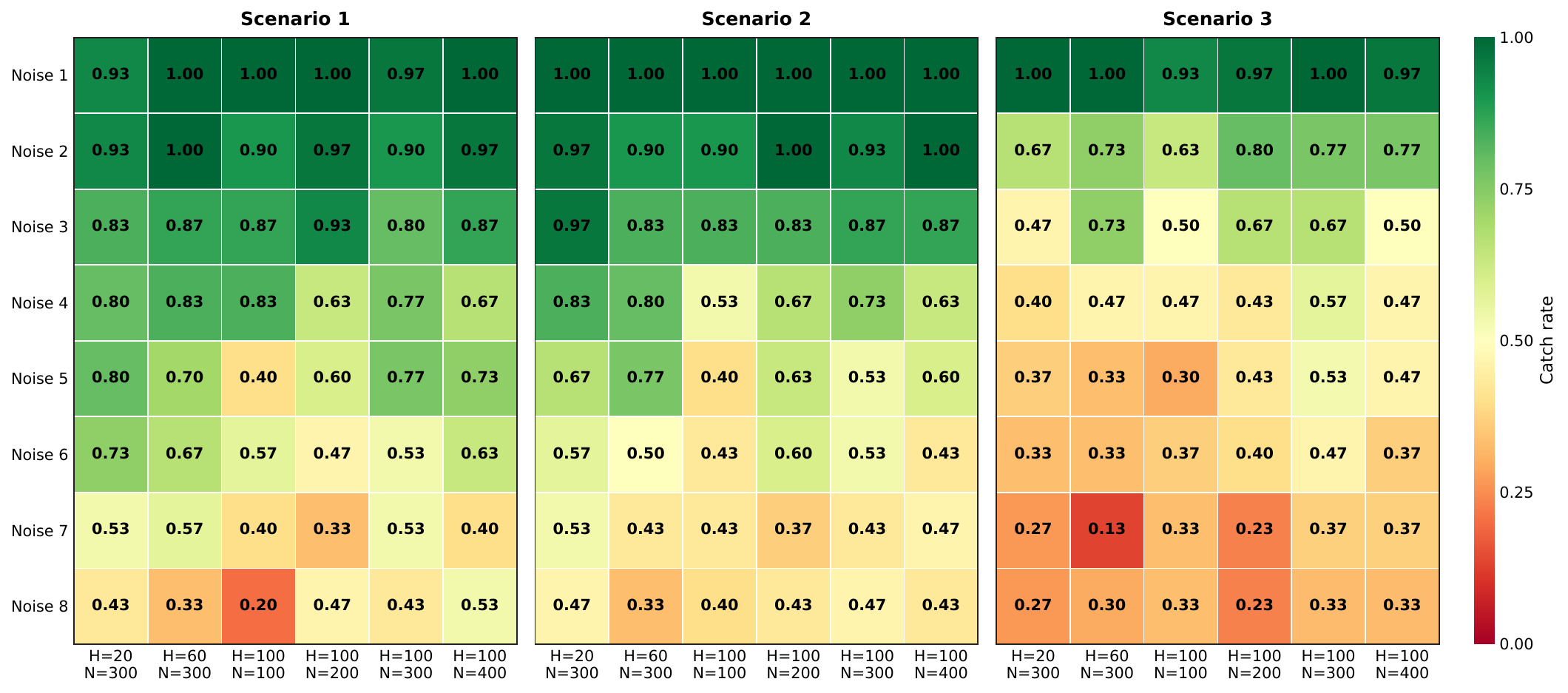}
 \caption{\textbf{Experiment 5.} State AIF performance heatmaps across all sampled plan variations ($N=\{100, \,200, \,300, \,400\}$) and planning horizons ($H=\{20, \,60, \,100\}$). The agent here has an initial belief covariance of the noise similar to the actual noise conditions listed in table \ref{tab:noise-state}.}
\label{fig:agent_knows}
\end{figure}

Our first set of experiments evaluated the active inference agent in cases where the agent does not have the true model of the observation noise. We first performed a preliminary experiment to assess the effect of bat size on task difficulty. Since the bat length $L$ determines the margin of error tolerated at interception of the ball by the bat, varying $L$ allowed us to calibrate the difficulty of the benchmark. Based on these experiments we fixed $L = 0.0625$, the value at which the task remained reliably solvable while still imposing a non-trivial control and estimation demand. Figure \ref{fig:batsize} summarises our results for the three different bat sizes assessed. \\
We next examined how performance scales with the breadth of the planning search. Holding the horizon fixed at $H = 100$, we swept the number of sampled plans $N \in {100, 200, 300, 400}$. Since the agent selects actions by minimising expected free energy over $N$ sampled plan rollouts, $N$ determines how thoroughly the space of all possible future actions across different planning trajectories is explored at each step. Figure \ref{fig:nplans} summarises our results for the planning sweeps. As expected, increasing the number of plans provides the agent with the ability to search across a larger space of possible candidate actions to take, compensating for the increased uncertainty with growing noise. We observed $N=300$ to be the optimal trade-off between computation time and successfully completed trials where the bat intercepted the ball across all evaluated scenarios. \\
We varied the planning horizon $H \in {20, 60, 100}$ while holding the number of randomly sampled plans fixed at $N = 300$. The horizon sets how many steps into the future each candidate policy is rolled out under the generative model before its expected free energy is evaluated. We observed that a longer horizon allowed the agent to anticipate the ball's trajectory earlier and position the bat in advance, whereas a shorter horizon yielded worse catch rates, especially in the harder scenario (scenario 3) where the agent needed to infer the ball velocity as well as its lateral trajectory. The catch rates are reported as a function of $H$ across all three scenarios and eight observation-noise conditions in the figure \ref{fig:horizons} heatmaps. We provide the deterministic baseline as a reference for our results. As the deterministic baseline differs from the active inference agents in its planning, it allows us to assess the true performance of these agents under uncertainty. Particularly in cases of high uncertainty, as the scenarios get progressively harder, the catch rate for the bat intercepting the ball in deterministic baseline deteriorate, while the state AIF agent with access to longer planning horizons and sufficient planning reliably discards noisy observations to choose actions to minimise its uncertainty about the environment to intercept the ball. \\
We also evaluated the effect of grid sizes on the pixel AIF agent. Our results are summarised in figure \ref{fig:gridsize}. We observed that an increase in the grid size resulted in a worse catch rate with increasing noise. This is due to the inability of the neural network encoder to handle out of distribution pixel perturbations. The neural network encoder (MLP) here serves are the measurement model of the agent compressing the pixel grid into its 4D observation vector. Increasing the noise in the pixel space results in increasingly accumulating errors leading to poor observation vector estimates which resulted in poorer performance. Due to this we used a $32 \times 32$ grid size for our experiments as it provided the best trade off between catch rates and the need to compute multiple forward passes of higher dimensional grid sizes through the measurement model to obtain the 4D observations. \\
In our final experiment (figure \ref{fig:agent_knows}), we evaluated how an agent whose noise model is calibrated to the true noise in the observations performs for the state AIF agent. Even in cases of higher uncertainty, the agent continues to act in accordance with its preference distribution, appropriately discarding unreliable observations rather than being driven by them.

\section{Discussion}

\paragraph{\textbf{Plan Sensitivity.}}

\begin{figure}[htbp]
 \centering
 \includegraphics[width=1\textwidth]{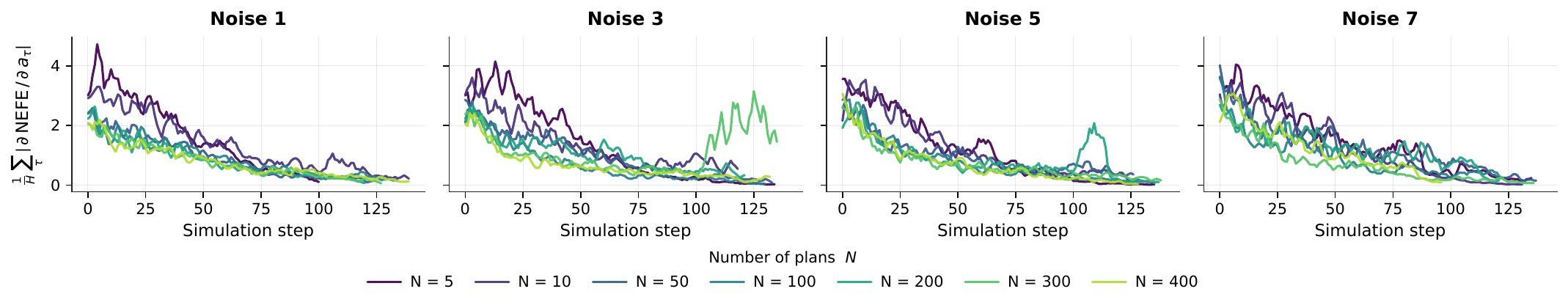}
 \caption{Sensitivity analysis of the agent's chosen plan for four example noise conditions by minimizing the NEFE with respect to all actions in its planning horizon. The plot shows the partial derivative of the actions in the plan with respect to the NEFE. An increase in the number of sampled plans decreases the gradient such that a small perturbation in the actions does not change the agent's chosen plan's NEFE.}
\label{fig:gradient}
\end{figure}

\begin{figure}[htbp]
 \centering
 \includegraphics[width=1\textwidth]{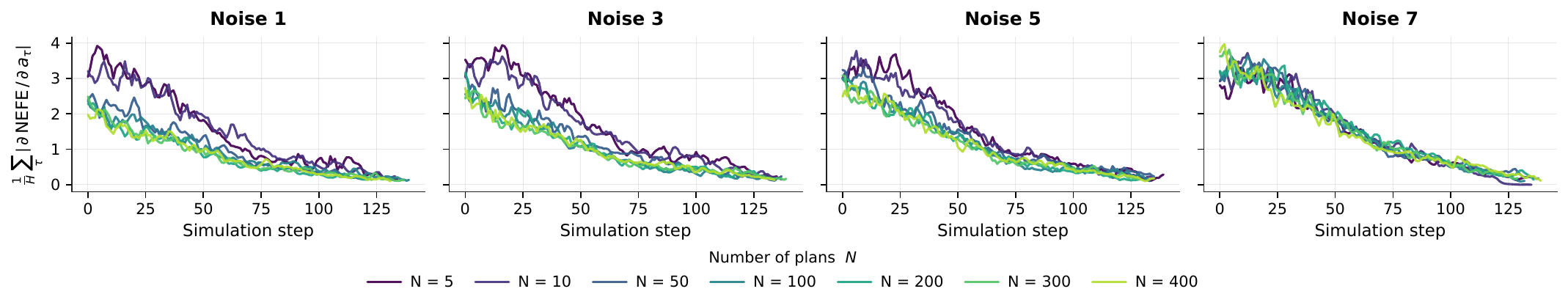}
 \caption{Sensitivity analysis of the agents planning for four example noise conditions across different number of plans. The plot shows the state AIF agent's sensitivity to the plan chosen by minimizing the NEFE to all the actions in its planning horizon for the case where the agent's belief of the noise matches the actual noise. An increase in the number of sampled plans decreases the gradient.}
\label{fig:gradient_agent_knows}
\end{figure}

Figures \ref{fig:gradient} and \ref{fig:gradient_agent_knows} report the sensitivity of the state AIF agent's selected plan, measured as the average of the partial derivative of the negative expected free energy (NEFE) with respect to each action in the planning horizon (for four noise conditions from table\ref{tab:noise-state}). A large gradient magnitude indicates that the chosen plan lies on a steep region of the NEFE landscape, such that small perturbations to the actions would substantially change the plan's score. A small gradient indicates that the plan sits near a flat optimum and is correspondingly robust to such perturbations. We sampled $\mathrm{N} \in \{5, 10, 50, 100, 200, 300, 400\}$ plans across a $H=100$ planning horizon and averaged the gradient over five runs for each plan and each noise condition evaluated. We observe that as the number of sampled plans increases, the average gradient magnitude across the horizon decreases. With more candidate plans evaluated at each step, the agent is more likely to recover a plan close to the true NEFE minimum. Increasing the number of random plans drives the agent toward solutions that not only lead to better action selection but are also less sensitive to the precise choice of actions. This trend holds across all four noise conditions shown, and is preserved in the noise-aware case (Figure~\ref{fig:gradient_agent_knows}), where calibrating the agent's belief to the true observation noise yields the same flattening of sensitivity as the number of plans grows.

\paragraph{\textbf{Environment Uncertainty.}}

\begin{figure}[htbp]
 \centering
 \includegraphics[width=0.9\textwidth]{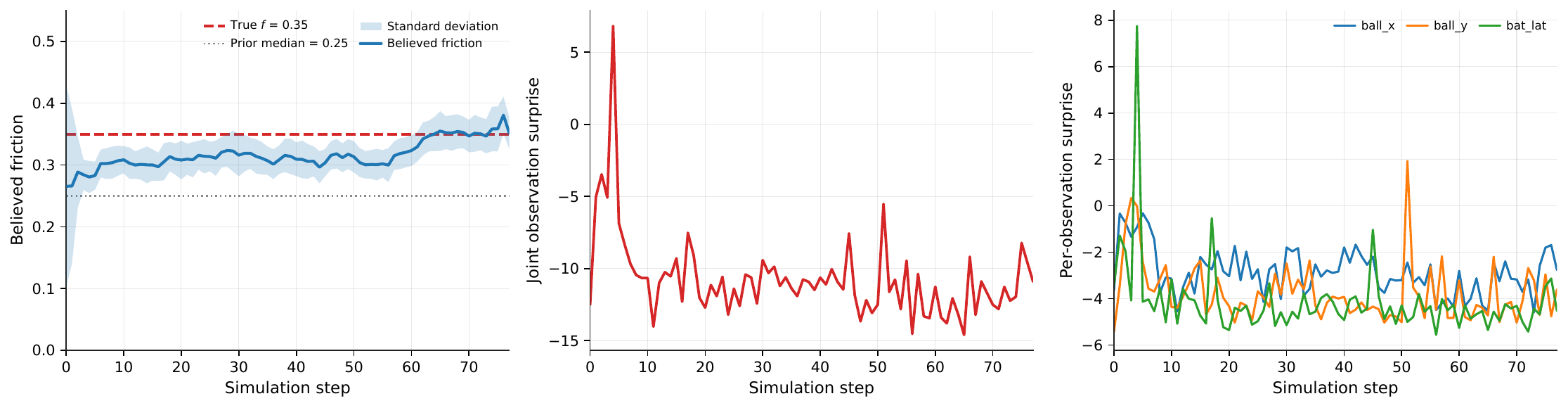}
 \caption{(Left) Evolution of the state AIF agent's belief about the friction field around the bat vs the true friction field. (Middle) The agent's surprise over its observation vector (Right) Agent's surprise over balls' $x$ and $y$ position and the bat's lateral position.}
\label{fig:friction1d_plots}
\end{figure}

Our experiments do not explicitly evaluate the integration of the information gain term in the free energy computation. This is because the agents do not need an epistemic drive to explore the environment as the system dynamics are known and observable. To demonstrate the contribution of the epistemic term, we create a setting in which a relevant state component is not directly observable, but must be inferred. We do this by adding a friction field across the bat's lateral axis such that each action the agent takes is now damped by this field. We ran a state AIF agent to then assess whether the agent can infer this unknown field in the Noise 1 setting for scenario 3 (lateral falling ball with varying velocity). To run this agent, we scale the pragmatic value by a factor of $\beta=5$ and clip the information term to have a maximum value of $1000$. The final negative expected free energy is a sum of both terms. The agent updates its belief by computing the VFE for $100$ SGD steps with a learning rate of $10^{-3}$ after each new observation. The information-gain term in the expected free energy is approximated by sampling, $n_{\mathrm{IG}}^{s} = 9$ latent-state samples and, for each, $n_{\mathrm{IG}}^{o} = 1$ observation sample. Figure \ref{fig:friction1d_plots} shows how the agent's belief of the friction field changes over the simulation as it converges to the true value. We also show the agent's surprise of its observation for the entire 4D observation vector, as well as the individual components in the vector by computing the marginal probability of each component with respect to the agent's belief of the states. At the start of the simulation, the agent has a belief of $0.25$ on the friction field (true field = $0.35$). After the first few actions in the simulation we see a sharp rise in the agent's surprise over where the bat is as in a continuous state system, the surprise accumulates over a few time steps. Following this, the agent samples plans to minimise its uncertainty about the field as we see the friction field belief rise over the next few steps. This shows the agent isn't just minimising the EFE to satisfy a preference, but also has an epistemic drive to learn its environment. In future work, we aim to extend this scenario by introducing multiple friction-field bands across the bat's lateral axis to see how the agent can cope with varying regions of high and low friction affecting its actions.

\section{Conclusion}

In this paper, we presented \texttt{JAXbounce}, a continuous-control, Pong-like benchmark that moves active inference beyond the low-dimensional, discrete environments on which it is typically evaluated. The environment combines continuous bat dynamics and a continuous action interface with a calibrated family of observation-noise models, and is fully differentiable and parallelisable in JAX, making evaluating over an agent's planning hyperparameters cheap to run. We evaluate a state-based and a pixel-based active inference agent on \texttt{JAXbounce} across three increasingly difficult interception scenarios and eight observation-noise conditions, characterising how its performance scales with the number of plans, its planning horizon, and knowledge of the noise process. We show that agents evaluated in our environment exhibit goal-directed behaviour even under high perceptual uncertainty. \texttt{JAXbounce} and all experiment configurations are open-sourced to the community with hopes to accelerate continuous-control active inference research.

\subsubsection*{Acknowledgements.}
This work was funded by the Designing Interaction Freedom via Active Inference (DIFAI) ERC Advanced Grant (101097708), funded under the UK Horizon guarantee as EPSRC EP/Y029178/1. R.M-S. and J.H.W. acknowledge EPSRC EP/R018634/1. R.M-S. also acknowledges EP/T00097X/1 and EP/T021020/1. M.K. acknowledges the AI-Twin ERC Advanced Grant (101200584), funded by HORIZON.1.1--European Research Council (ERC).

\newpage

\appendix

\section{Pixel-based Encoder Pretraining}
\label{sec:encoder}

The pixel-based AIF encoder is pretrained offline on $30000$ random (state, image) pairs by minimising $\mathbb{E}_{\mathbf{x},\eta}\Vert\phi_\theta(\eta(\mathrm{render}(\mathbf{x}))) - \mathbf{o}^{\mathrm{s}}(\mathbf{x})\Vert_2^{2}$, where $\eta$ is noise drawn uniformly from one of four categories, (1) clean, (2) additive Gaussian (std $\sim\mathcal{U}[0,0.02]$), (3) Hann-window blur (window $2 \times 2$), and (4) salt-and-pepper (per-channel rate $\sim\mathcal{U}[0,0.005]$). We do this to test the effects of having a high noise in input measurements in our simulations. Neural network performance tends to degrade drastically in the presence of high noise and adversarial attacks (new noise). Due to this reason, we make sure that the encoder is familiar with each noise structure but has never seen the test-time noise distribution, creating a controlled out-of-distribution gap. All MLPs are trained using the Adam optimiser with a learning rate of $1e^{-2}$ for $8000$ steps with batch size $128$.

\section{Noise Conditions}

We provide the different noise conditions applied to the deterministic agent and the state AIF agent in tabular form in table \ref{tab:noise-state} below. Figure \ref{fig:noise_examples} visualizes the different noise conditions applied to the pixel grid for the pixel AIF agent. 

\begin{figure}[htbp]
 \centering
 \includegraphics[width=0.8\textwidth]{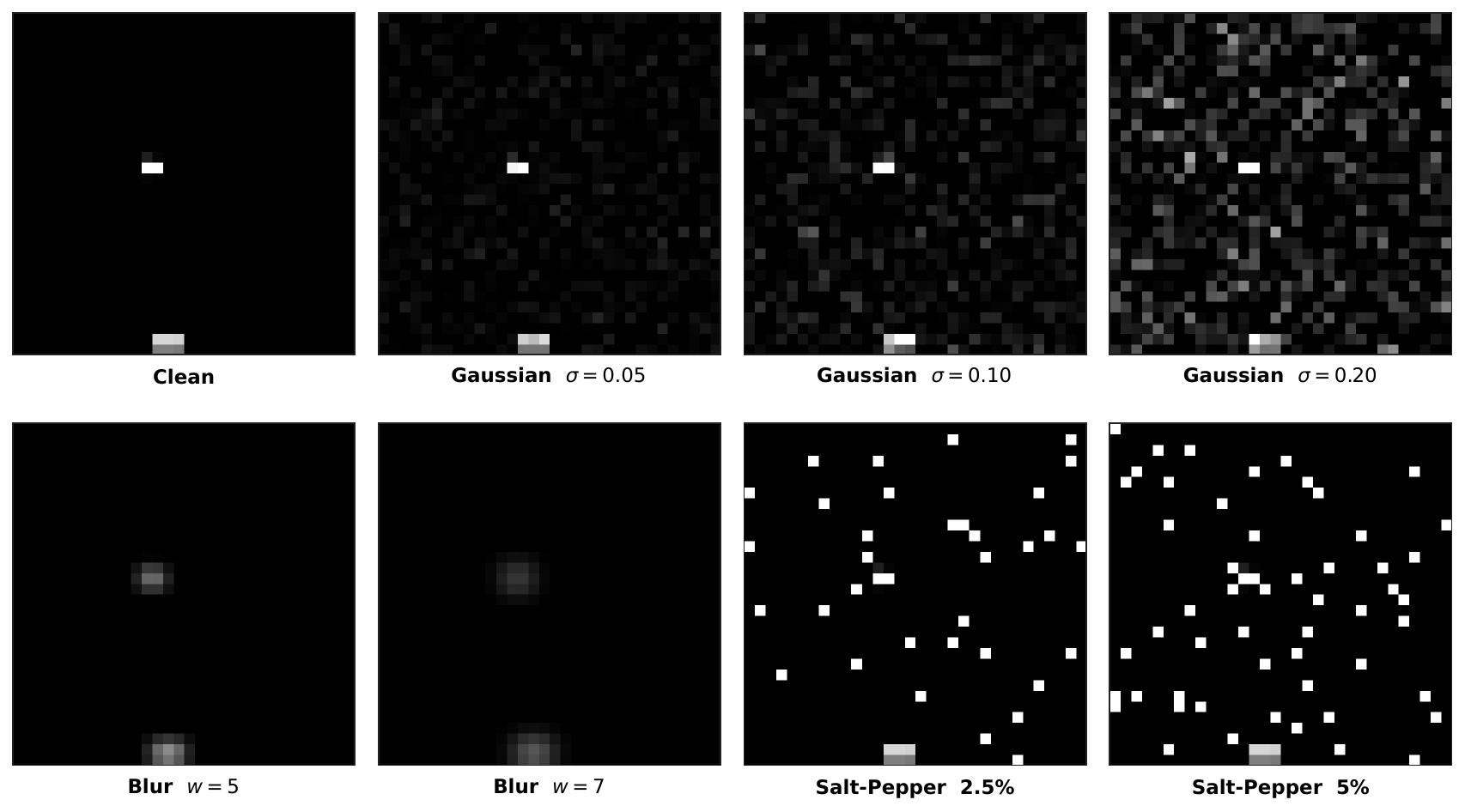}
 \caption{Visualisations of the \texttt{JAXbounce} grids used in pixel AIF simulations. The clean grid with increasing uncertainty on the agent's initial belief (table \ref{tab:noise-state}) of the position of the ball and the bat and increasing uncertainty on its measurement model is used for the deterministic and state AIF simulations.}
\label{fig:noise_examples}
\end{figure}

\begin{table}[h]
\centering
\small
\caption{Noise conditions in experiments for the deterministic baseline and state-based AIF agent. The eight noise tiers are designed to make inference increasingly harder for the agents.}
\label{tab:noise-state}
\begin{tabular}{l|ccc}
\toprule
Label & $\sigma_{\mathrm{pos}}^{(0)}$ & $\sigma_{\mathrm{v}}^{\mathrm{mult}}$ & $\sigma_{\mathrm{o}}^{\mathrm{inj}}$ \\
\midrule
\textsc{Noise 1}      & $10^{-3}$ & $1.0$ & $10^{-3}$ \\
\textsc{Noise 2}  & $0.02$ & $1.5$ & $0.05$ \\
\textsc{Noise 3}  & $0.05$ & $2.0$ & $0.10$ \\
\textsc{Noise 4} & $0.10$ & $2.5$ & $0.20$ \\
\textsc{Noise 5}     & $0.12$ & $3.0$ & $0.25$ \\
\textsc{Noise 6}     & $0.15$ & $3.5$ & $0.30$ \\
\textsc{Noise 7}     & $0.18$ & $4.0$ & $0.35$ \\
\textsc{Noise 8}    & $0.25$ & $5.0$ & $0.40$ \\
\bottomrule
\end{tabular}
\end{table}

\section{Results Heatmaps}

\begin{figure}[htbp]
 \centering
 \includegraphics[width=1.0\textwidth]{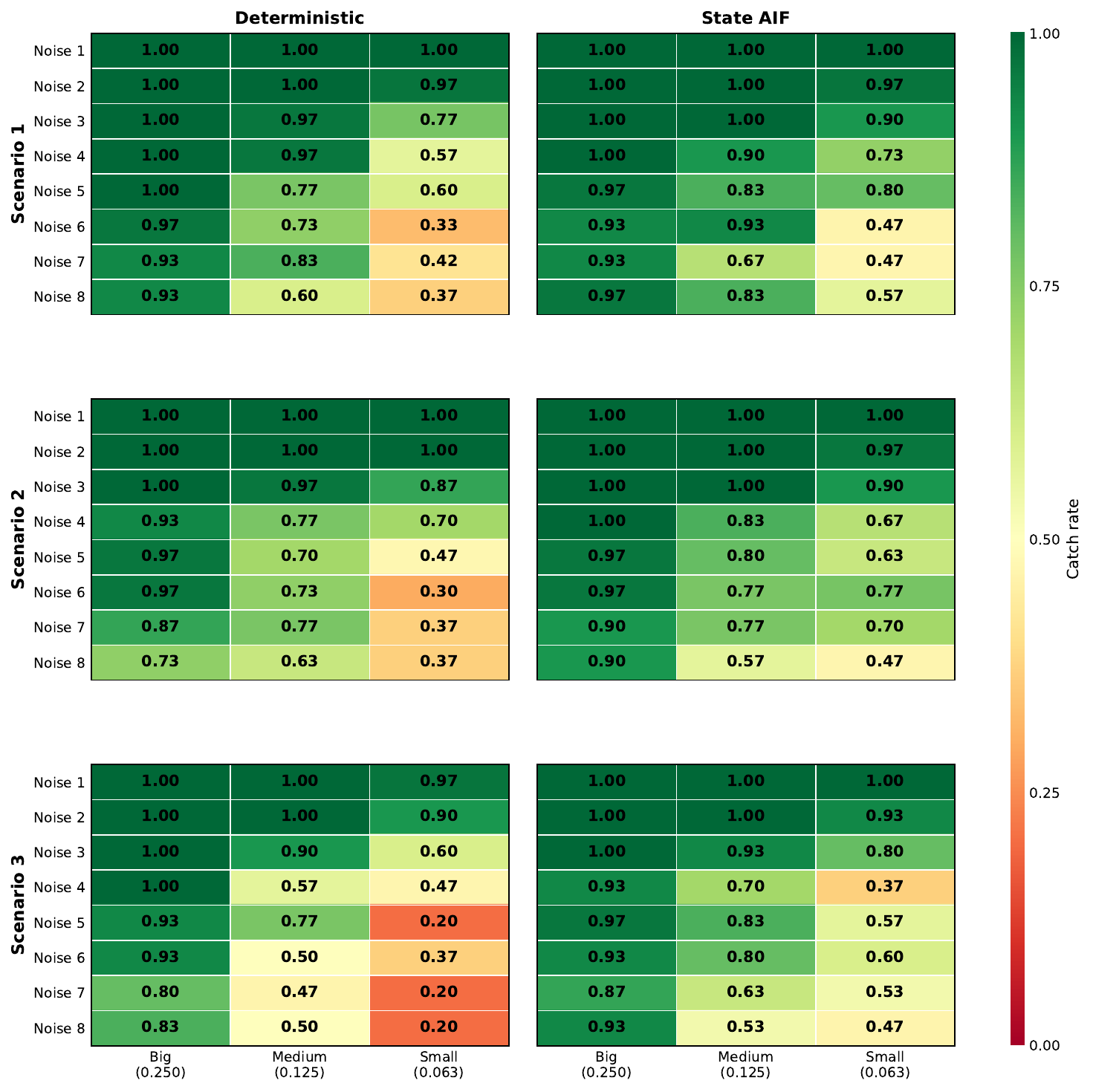}
 \caption{\textbf{Experiment 1.} Bat size comparison heatmaps between the deterministic and state AIF agents. All runs are averaged over 30 trials for Scenario 1 (constant velocity ball falling), Scenario 2 (variable ball velocity in each run) and Scenario 3 (variable ball velocity and lateral ball falling). State AIF agent runs randomly sample 300 plans over a finite horizon of 100 timesteps.}
\label{fig:batsize}
\end{figure}

\begin{figure}[htbp]
 \centering
 \includegraphics[width=1.0\textwidth]{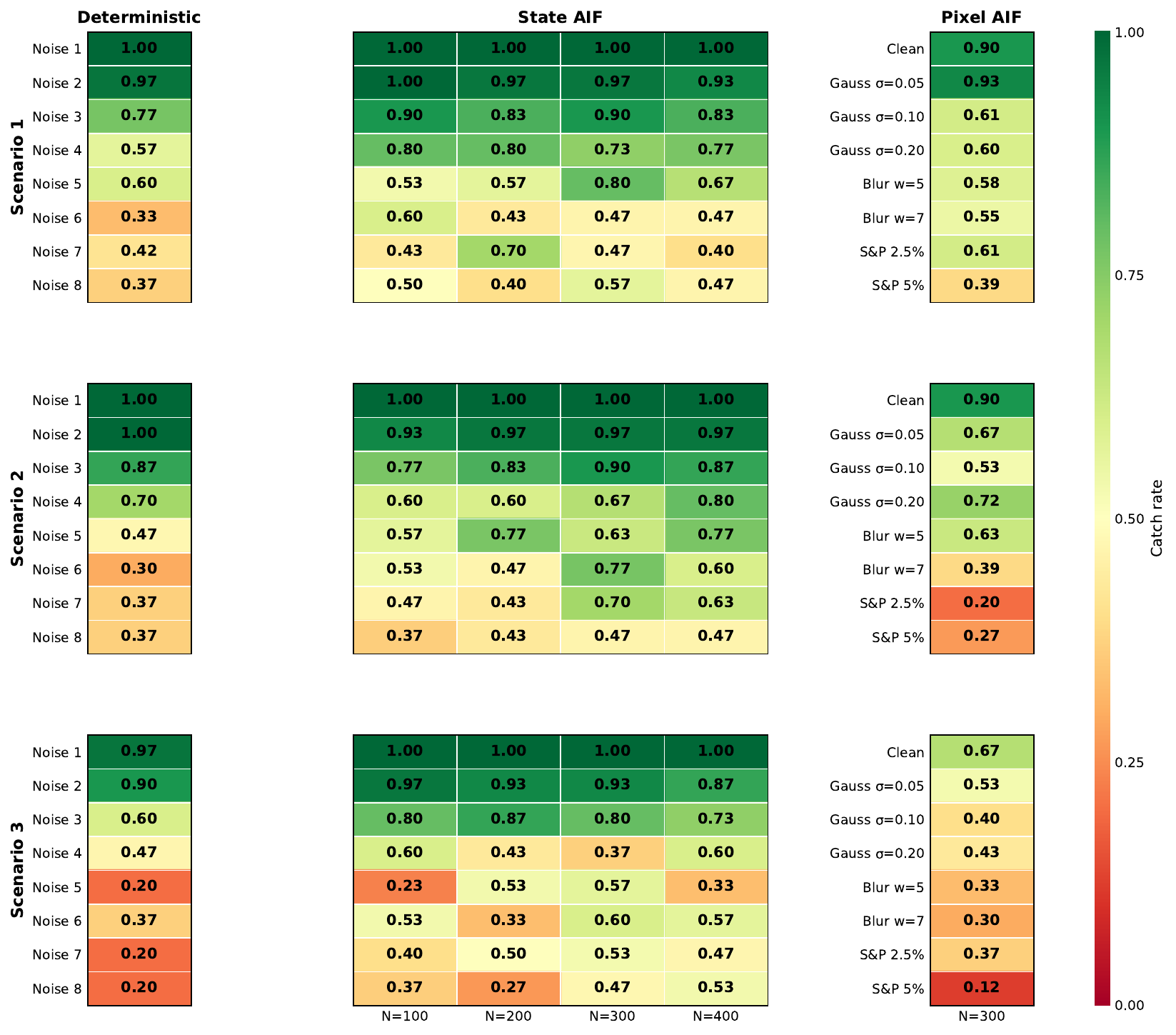}
 \caption{\textbf{Experiment 2.} Analysis of agent performance for varying number of plans across the three scenarios. All plans are run for a horizon of 100.}
\label{fig:nplans}
\end{figure}

\begin{figure}[htbp]
 \centering
 \includegraphics[width=1.0\textwidth]{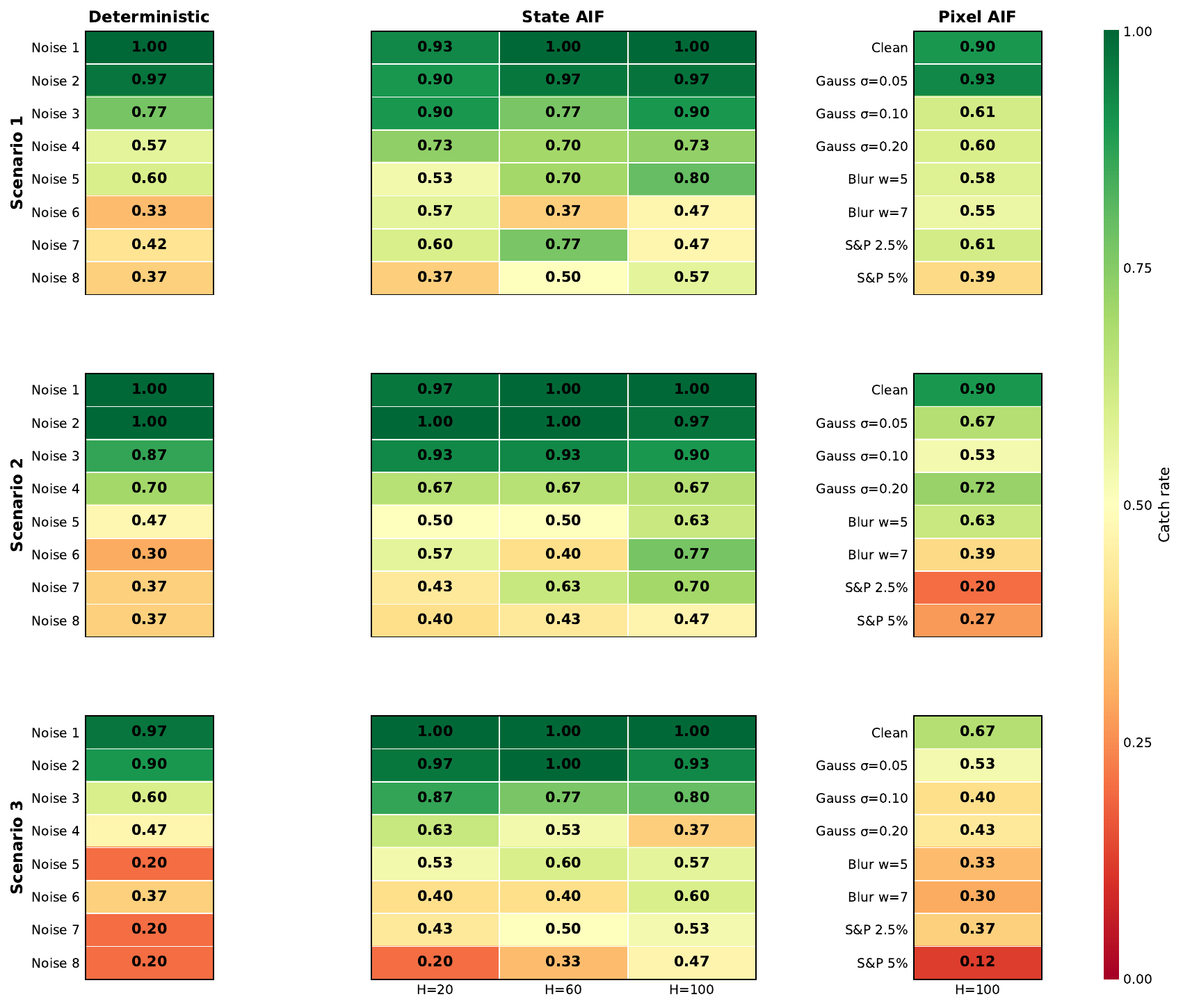}
 \caption{\textbf{Experiment 3.} Comparison of performance for varying planning horizons across the three scenarios. The agents randomly sample 300 plans.}
\label{fig:horizons}
\end{figure}

\begin{figure}[htbp]
 \centering
 \includegraphics[width=1.0\textwidth]{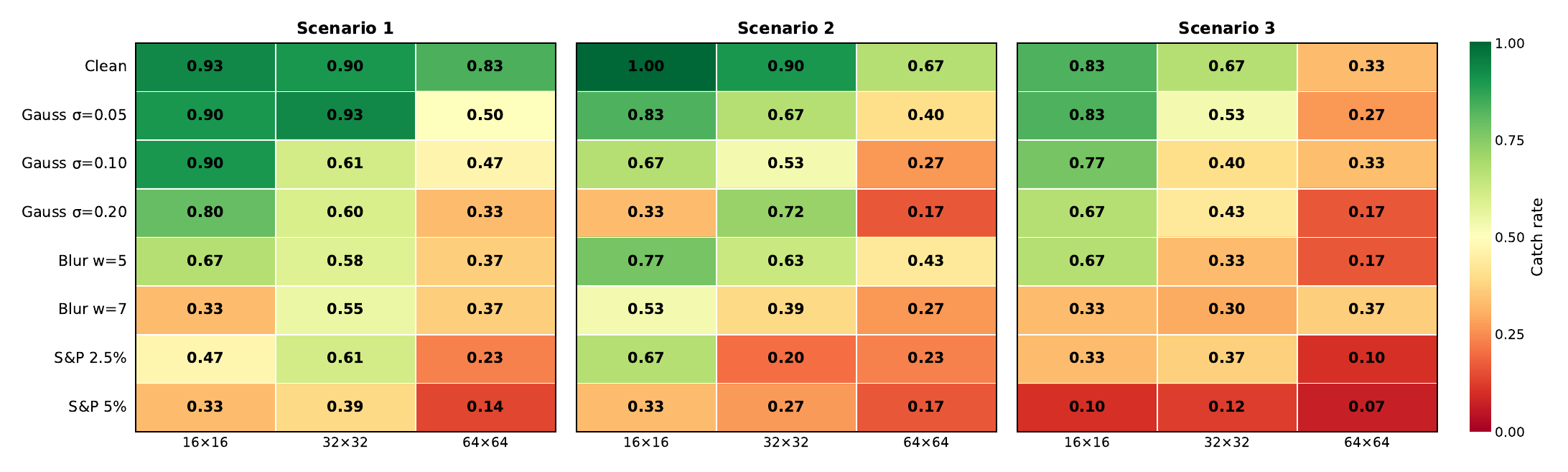}
 \caption{\textbf{Experiment 4.} Varying grid sizes for the pixel AIF agent. Each simulation is run for 300 randomly sampled plans over a planning horizon of 100 time steps.}
\label{fig:gridsize}
\end{figure}

We visualise all catch rate heatmaps for experiment 1 (figure \ref{fig:batsize}), experiment 2 (figure \ref{fig:nplans}), experiment 3 (figure \ref{fig:horizons}) and experiment 4 (figure \ref{fig:gridsize}) in this section.

\section{Limitations} 

A single agent for configuration ($N = 300$ policies, $H = 100$ horizon, $T = 150$ steps) performs roughly $4.5$ million forward passes through its generative model per trial. Our evaluations report over $\mathbf{10,000}$ agent simulations. Across all agents, we have run over $\mathbf{20,000}$ simulations with different seeds in our initial experiments. This requires considerable amount of compute. For the pixel AIF agent specifically, we have run approximately $2.9$ billion forward passes of the MLP encoder parallelizing runs across $4$ Nvidia RTX A6000 GPUs for $3$ days. This is considerable amount of compute required to run these agents even with small MLP encoders which is a current limitation of our approach. We aim to mitigate this in future work by investigating how to sample plans in smarter and more efficient ways \cite{apers2024hamiltonian} as well as investigating ways to amortise \cite{huang2026aline} active inference for our pipeline.

\bibliographystyle{splncs04}
\bibliography{references}

\end{document}